\documentclass[runningheads]{llncs}

\usepackage{eccv}

\usepackage{eccvabbrv}

\usepackage{graphicx}
\usepackage{booktabs}

\usepackage[accsupp]{axessibility}  % Improves PDF readability for those with disabilities.

\usepackage{amsmath}
\newcommand{\Real}{\mathbb{R}}

\newcommand{\trans}{^{\intercal}}

\newcommand{\softmax}{\mathrm{softmax}}

\newcommand{\sigmaith}{\sigma_i}
\newcommand{\sith}{s_i}
\newcommand{\xith}{x_i}
\newcommand{\xquery}{x_q}
\newcommand{\fith}{f_i}

\newcommand{\frefine}{f^{\mathrm{(refine)}}}

\newcommand{\Qbf}{\mathbf{Q}}
\newcommand{\Kbf}{\mathbf{K}}
\newcommand{\Vbf}{\mathbf{V}}

\usepackage{enumitem}
\setlist[itemize]{parsep=0pt,partopsep=0pt,leftmargin=*,itemsep=5pt}
\setlist[enumerate]{parsep=0pt,partopsep=0pt,leftmargin=*,itemsep=5pt}
\makeatletter
\renewcommand\paragraph{\@startsection{paragraph}{4}{\z@}%
    {.5em}%
    {-1em}%
    {\normalfont\normalsize\bfseries}}
\makeatother
\usepackage{hyperref}

\usepackage{orcidlink}

\begin{document}

\title{ASSR-NeRF: Arbitrary-Scale Super-Resolution on Voxel Grid for High-Quality Radiance Fields Reconstruction
}

\titlerunning{ASSR-NeRF}

\author{Ding-Jiun Huang\inst{1}\orcidlink{0000-1111-2222-3333} \and
Zi-Ting Chou\inst{1}\orcidlink{1111-2222-3333-4444} \and
Yu-Chiang Frank Wang\inst{1,2}\orcidlink{2222--3333-4444-5555} \and
Cheng Sun\inst{1,2}\orcidlink{2222--3333-4444-5555}}

\authorrunning{Huang et al.}

\institute{National Taiwan University \and
NVIDIA Research}

\maketitle
\begin{abstract}
NeRF-based methods reconstruct 3D scenes by building a radiance field with implicit or explicit representations. While NeRF-based methods can perform novel view synthesis (NVS) at arbitrary scale, the performance in high-resolution novel view synthesis (HRNVS) with low-resolution (LR) optimization often results in oversmoothing. On the other hand, single-image super-resolution (SR) aims to enhance LR images to HR counterparts but lacks multi-view consistency. To address these challenges, we propose Arbitrary-Scale Super-Resolution NeRF (ASSR-NeRF), a novel framework for super-resolution novel view synthesis (SRNVS). We propose an attention-based VoxelGridSR model to directly perform 3D super-resolution (SR) on the optimized volume. Our model is trained on diverse scenes to ensure generalizability. For unseen scenes trained with LR views, we then can directly apply our VoxelGridSR to further refine the volume and achieve multi-view consistent SR. We demonstrate quantitative and qualitatively that the proposed method achieves significant performance in SRNVS.
\keywords{neural radiance field \and super-resolution \and feature distillation}
\end{abstract}   
\section{Introduction}
\label{sec:intro}

\begin{figure*}[t]
	\centering
	\includegraphics[width=1\columnwidth]{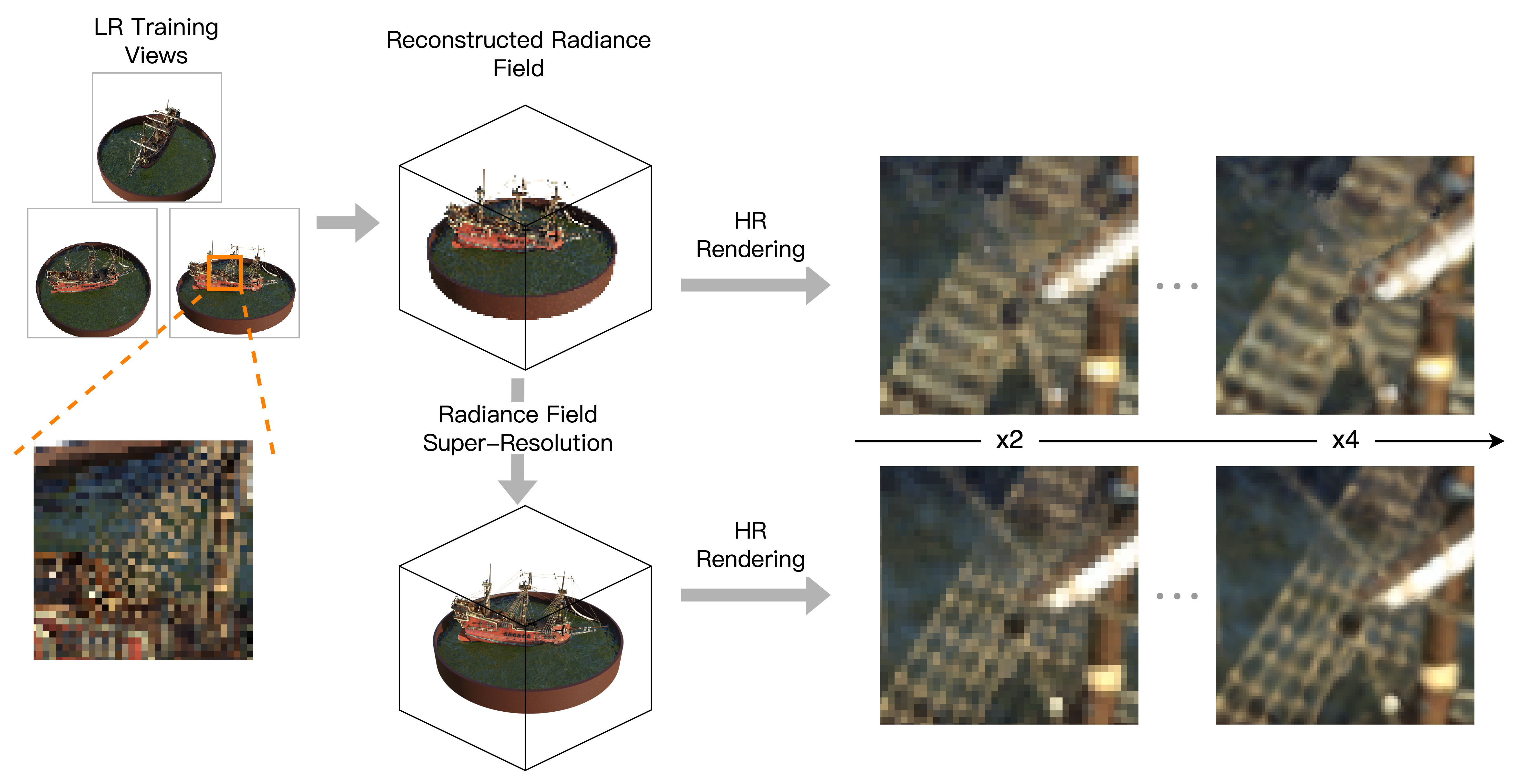}
 % \vspace{-5mm}
    \caption{ 
        Given a radiance field reconstructed from low-resolution (LR) training views, we perform radiance field super-resolution, leading to cleaner details in rendered views of high-resolution (HR).
    }
    % \vspace{-3mm}
	\label{fig:teaser}
\end{figure*}

Novel view synthesis (NVS), or 3D scene reconstruction, aims to synthesize images of a 3D scene from arbitrary viewing directions given multi-view images and camera poses.  NeRF~\cite{mildenhall2020nerf} achieves remarkable NVS results by employing neural network as an implicit volumetric representation, which maps 3D positions and viewing directions to view-dependent colors and occupancy.
% Since NeRF learns a continuous volumetric representation, it is capable of synthesizing novel views at arbitrary resolution.
% Although NeRF can generate appealing results, it has lengthy training and rendering time. Many following works~\cite{sun2022direct, yu2021plenoxels, chen2022tensorf, M_ller_2022, barron2023zipnerf, reiser2021kilonerf, yu2021plenoctrees} then put emphasis on reducing training and rendering time. For example, Intant-ngp~\cite{M_ller_2022} takes only minutes to reconstruct a 3D scene. Another issue of NeRF is aliasing, happening when training views and rendering views are different in resolutions. Methods ~\cite{barron2021mipnerf, barron2023zipnerf, hu2023trimiprf} then propose to mitigate this issue with mipmap techniques.
Due to its flexibility, numerous follow-up extensions, applications, and improvements had been made on top of NeRF.
While current state-of-the-art NeRF-based methods can accurately synthesize geometry and appearance of a scene, high-resolution novel view synthesis (HRNVS) poses a great challenge to them, where high-resolution (HR) novel views are rendered by radiance fields constructed from low-resolution (LR) training views. Since LR training views lack details of a scene, the rendered HR novel views are blurry and noisy.

On the other hand, single-image super-resolution (SISR) aims to synthesize an HR image from its LR counterpart. Different from mathematical up-sampling methods, e.g., bilinear interpolation, SISR methods~\cite{Dong2014ImageSU, Lim2017EnhancedDR, Zhang2018ImageSU, Ledig2016PhotoRealisticSI, Liang2021SwinIRIR, Lu2021TransformerFS, Wang2018ESRGANES} integrate deep learning models to enrich details and textures that are missed in LR images. Recently, generative-based methods~\cite{wang2023exploiting, Li2021SRDiffSI} shows exciting results with the advent of diffusion models.

A straightforward way of solving the above-mentioned quality issue of HRNVS is to directly apply SISR methods on the rendered views. However, applying SISR on each view independently will cause multi-view inconsistency, i.e., the geometry or appearance of objects isn't consistent among the multiple rendered views.~\cite{Wang_2022} first proposes NeRF-SR for super-resolution (SR) of neural radiance field. Given a set of LR training views and 1 HR reference view of the same scene, NeRF-SR refines the rendered HR view through super-sampling and a patch-based refinement module. Although it shows satisfying results, requiring an HR reference view for every scene is not practical. With similar ideas, Super-NeRF~\cite{han2023supernerf} and CROP~\cite{10205145} propose to employ an SR module to guide the HR renderings of NeRF, and rendered novel views can be recycled to guide the SR module, making its SR outputs view-consistent. However, a lengthy optimization is required for every scene, or the upscaling factor of SR is fixed, reducing the flexibility. A pre-print~\cite{bahat2023neural} proposes to decompose a neural radiance field to tri-plane~\cite{chan2022efficient}, and applies a pre-trained SISR model to the 2D feature planes. While this design improves the generalizability of SR module, i.e., a trained or fine-tuned SR module in a scene can be directly applied to another scene, applying SR on tri-plane independently causes inconsistency between the planes. 

In this work, we propose arbitray-scale super-resolution NeRF (ASSR-NeRF) for super-resolution novel view synthesis (SRNVS) without the above-mentioned issues. While HRNVS, performed by common NeRF-based models, only raise the resolution of rendered views without adding more details, SRNVS aims to enrich the details and textures in novel views and remains multi-view consistency. ASSR-NeRF consists of two main parts: a voxel-based distilled feature field reconstructed from LR views and an attention-based VoxelGridSR model that will directly super-resolve the radiance field. Inspired by~\cite{sun2022direct, yu2021plenoxels, yu2021plenoctrees}, we first construct a distilled feature field to represent a scene with explicit voxel grids. To let VoxelGridSR perform attention on meaningful features, feature distillation is deployed to embed extracted features of 2D training views into 3D voxel grids. Since our SR is performed in 3D space, there won't be multi-view inconsistency. The design of VoxelGridSR is inspired by LIIF~\cite{chen2021learning}, which treats SR as a mapping problem between coordinates and the corresponding RGB values. Given the coordinate of a queried point, VoxelGridSR performs $\textit{density-distance-aware attention}$ on distilled features queried from its local region and outputs a refined feature representing the queried point. Since the coordinate in 3D space is continuous, VoxelGridSR is capable of optimizing SR at arbitrary scale. In addition, since VoxelGridSR is generalizable, it serves as an off-the-shelf method that can directly apply to any reconstructed feature field of unseen scene for SRNVS.

In summary, our key contributions of our work are as follows:
\begin{itemize}
    \item We propose a novel framework, ASSR-NeRF, for super-resolution novel view synthesis (SRNVS) of radiance field.
    \item We distilled knowledge of low-level 2D SR priors from pre-trained feature extractor into radiance field to benefit SR in 3D space 
    \item We design VoxelGridSR model to refine the optimized volume for SRNVS with richer textures and details.
    \item We train our VoxelGridSR to be generalizable so we can directly refine the radiance field of unseen scenes trained with LR views.
\end{itemize}
\section{Related Work}
\label{sec:related}

\subsection{Image Super-Resolution}

Image super-resolution (SR) aims to restore a high-resolution (HR) image from its low-resolution (LR) counterpart. Early image SR methods~\cite{Dong2014ImageSU, Lim2017EnhancedDR, Zhang2018ImageSU, Ledig2016PhotoRealisticSI} adopt a deep convolutional neural network (CNN) to improve performance. After the advent of the attention mechanism, methods such as SwinIR~\cite{Liang2021SwinIRIR} and ESRT~\cite{Lu2021TransformerFS} achieve competitive performance using a transformer-based architecture. To further enrich details in SR results, GAN-based and diffusion based methods~\cite{Wang2018ESRGANES, Li2021SRDiffSI, wang2023exploiting} generate finer details as well as rich textures through adversarial training and powerful diffusion models respectively. Although excelling in image SR tasks,  most methods can only perform SR on one fixed scale, failing to fit in real-world scenarios where display devices come in different resolutions. To perform arbitrary-scale super-resolution (ASSR), one could first properly upscale the input image, then apply existing image SR methods. However, this approach is time-consuming and would lead to unsatisfied results when the upscaling factor is too large. Recently, several methods~\cite{hu2019metasr, chen2021learning, lee2022local, wei2023superresolution, cao2023ciaosr, yao2023local} are proposed to tackle ASSR with a single model. LIIF~\cite{chen2021learning} maps arbitrary coordinates to RGB colors with an MLP, taking encoded image latent as input. With the same idea as LIIF, CiaoSR~\cite{cao2023ciaosr} further applies attention mechanisms for an enlarged receptive field and ensemble of local predictions.

\subsection{Neural Radiance Fields}

NeRF~\cite{mildenhall2020nerf} has emerged as a prominent method for novel view synthesis (NVS), showcasing remarkable results with several input views and known camera poses. Specifically, NeRF encodes appearance and geometry of a 3D scene into a multi-layer perceptron (MLP), which takes 3D positions and viewing directions as input and predicts corresponding colors and densities. Volume rendering techniques then accumulate the queried properties along a camera ray to formulate the color of a pixel. Many follow-ups extend this idea to different settings and scenarios. Some methods dramatically improve training or rendering efficiency with explicit structures. DVGO and Plenoxel~\cite{sun2022direct, yu2021plenoxels} employ voxel grids as explicit scene representations, leading to fast convergence. TensoRF~\cite{chen2022tensorf} represents a scene with a tri-plane structure, greatly reducing both training time and memory usage. On the other hand, some methods focus on rendering quality. Mip-NeRF~\cite{barron2021mipnerf} leverages mipmapping to achieve anti-aliasing when rendering at different resolutions, and  Zip-NeRF~\cite{barron2023zipnerf} further integrates grid-based representations, inspired by Instant-ngp~\cite{M_ller_2022}, to enable both faster reconstruction and anti-aliased rendering, achieving state-of-the-art performance of NVS. 

\subsection{Super-Resolution of Neural Radiance Field}

Since NeRF~\cite{mildenhall2020nerf} learns a continuous volumetric representation for NVS, it can directly render novel views at arbitrary resolution. However, the rendering procedure adopted by NeRF samples a scene with a single ray per pixel, therefore producing renderings with aliasing, blurs or artifacts when training and rendering views vary in resolutions. Supersampling, which samples multiple rays per pixel, is an effective solution, but it leads to heavy computational burden for MLP queries. Applying existent image SR methods to rendered novel views is another straightforward approach. Nevertheless, super-resolving each view independently would cause multi-view inconsistency, i.e., geometry of an object in different views varies. Several methods~\cite{barron2021mipnerf, hu2023trimiprf, barron2023zipnerf} are proposed to mitigate this quality issue, but they  only “preserve” details, failing to “enrich” details that are missed in LR training views. For example, given LR training views of an antique vase, Mip-NeRF~\cite{barron2021mipnerf} can generate anti-aliased HR novel views but fail to restore finer patterns on the vase. NeRF-SR~\cite{Wang_2022} first proposes a module to refine details for rendered HR novel views with one HR reference view of the same scene. Following the same idea, RefSR-NeRF~\cite{10205402} performs reference-based SR and reaches massive speedup. ~\cite{10205145} further weakens the assumption that there’s always an HR reference image for each scene, proposing to super-resolve novel views with only LR training views. While these methods show impressive results, the SR modules are all trained with a fixed up-scaling factor or a per-scene optimization is required.

\section{Preliminaries}
\label{sec:preliminaries}

NeRF~\cite{mildenhall2020nerf} performs 3D scene reconstruction by encoding the geometry and occupancy of a scene into a multi-layer perceptron (MLP). The MLP maps a 3D position $x$ and a viewing-direction $d$ to the corresponding view-dependent color $c$ and density $\sigma$. NeRF marches ray to render the color $\mathbf{\hat{C}}(r)$ of each ray $r$ casting through a pixel. Along each ray, $K$ points are sampled to query the MLP for the corresponding color $c_i$ and density $\sigma_i$, which is then blended by:
% . $\mathbf{\hat{C}}(r)$ can then be obtained by the following equations: 
% \begin{equation} 
%     \hat{C}(r)=\sum_{i=1}^{K} T_i\, \alpha_i\, c_i, \;\; \alpha_i= 1 - exp(-\sigma_i\,\delta_i)
% \end{equation}
% \begin{equation}
%     T_i=\prod_{j=1}^{i-1}(1 - \alpha_j)
% \end{equation}
\begin{subequations}
\begin{align}
    \mathbf{\hat{C}}(r) &= \sum\nolimits_{i=1}^{K} T_i\, \alpha_i\, c_i ~, \\
    T_i &= \prod\nolimits_{j=1}^{i-1}(1 - \alpha_j) ~, \\
    \alpha_i &= 1 - \exp(-\sigma_i\,\delta_i) ~,
\end{align}
\end{subequations}
where $r$ are sampled rays; $T_i$ is the accumulated transmittance; $\alpha_i$ is the opacity; $(T_i\alpha_i)$ represents the probability of termination at point $i$; $\delta_i$ is the distance to adjacent points. NeRF model can then be trained with a photometric loss:
\begin{equation}
    L_{\mathrm{photo}} = \sum\nolimits_{r\in R}\| \mathbf{C}(r) - \mathbf{\hat{C}}(r)\|_2^2 ~.
\end{equation}

While NeRF shows appealing performance on novel view synthesis, it struggles with lengthy training and rendering time. 
Subsequent works~\cite{sun2022direct,yu2021plenoxels,M_ller_2022,chen2022tensorf} improve training efficiency by replacing the MLP with grid-based representations.
We build our super-resolution algorithm based on DVGO~\cite{sun2022direct}, where modalities of interest, e.g., density, color, of a 3D position are explicitly stored as voxel features and can be queried via trilinear interpolation:
\begin{equation}
    \mathrm{interp}(x, V):(\Real^3, \Real^{C\times N_x \times N_y \times N_z}) \rightarrow \Real^C
\end{equation}
where $V$ represents the voxel grid, $x$ is the 3D position, $C$ is the dimension of the modality, and $N_x,N_y,N_z$ represents the 3 dimensions of the grid respectively. We use a density grid for geometry and a feature grid for appearance.
A shallow MLP network, dubbed RGBNet, is additionally employed to map the queried voxel feature and the viewing-direction to view-dependent color. 

\section{Method}
\label{sec:method}
% \begin{figure*}[t]
% 	\centering
% 	\includegraphics[width=2\columnwidth]{figure/overview.png}
%  % \vspace{-2mm}
%     \caption{\textbf{Overview of ASSR-NeRF}: 
%     Given a query point $x^q$ along a ray, features and densities of the nearest $8$ neighbors are first sampled from voxel grids. Considering all the above modalities, VoxelGridSR then performs scaled dot-product attention for refined feature and density. Finally, view-dependent color is obtained through a cross-scene RGBNet.  
%     }
%     % \vspace{-3mm}
% 	\label{fig:overview}
% \end{figure*}

\begin{figure}[tb]
  \centering
  \includegraphics[width=1\columnwidth]{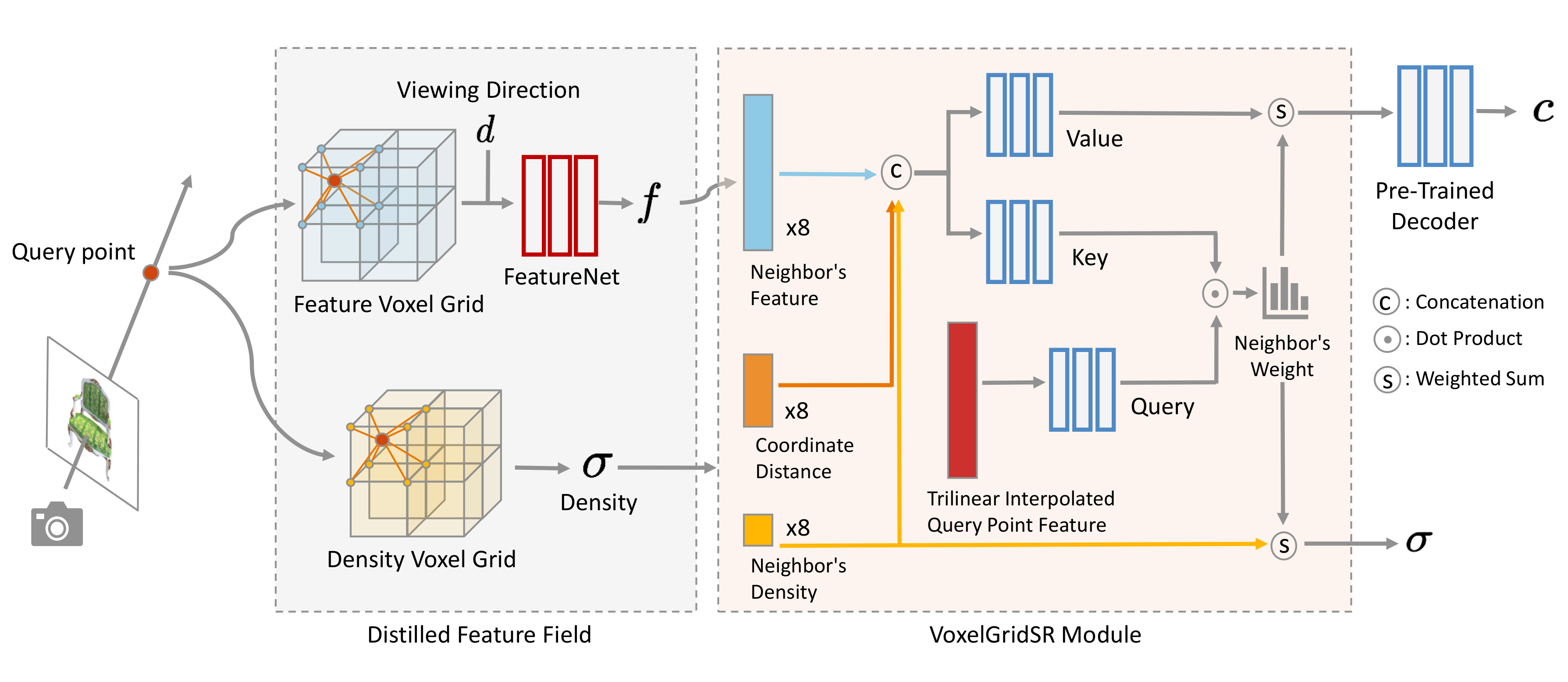}
  \caption{\textbf{Overview of ASSR-NeRF}: Given a query point $x$ along a ray, view-dependent distilled features and densities of its nearest neighbors are first sampled from a distilled feature field. Then, VoxelGridSR module aggregates the queried modalities and performs self-attention for refined feature and density. Finally, a pre-trained decoder maps the refined feature to RGB value $c$.
    }
  \label{fig:overview}
\end{figure}

\subsection{Overview}\label{sec:overview}
% In this section we describe our method for radiance field super-resolution. As shown in Fig.~\ref{fig:overview}, we propose a novel framework of arbitrary-scale super-resolution NeRF (\textbf{ASSR-NeRF}).
In this section we describe our method, dubbed ASSR-NeRF, for arbitrary-scale super-resolution NeRF. An overview of our approach is depicted in \cref{fig:overview}. ASSR-NeRF mainly consists of two parts: {\it (i)} a voxel-based distilled feature field and {\it (ii)} a generalizable VoxelGridSR module.
% For every scene, we first recontruct radiance field distilled with extracted features from 2D views so that VoxelGridSR model can utilize these SR priors and refine the optimized volume.
The distillation ensure the latent space alignment to facilitate multi-scene training and generalizability to novel scenes.
The VoxelGridSR learns to utilizes the distilled SR latent for radiance field refinement.
The following sections are organized as follows: we first describe our distilled feature field in Sec.~\ref{sec:dff} and explain why it is crucial in our approach. Then in Sec.~\ref{sec:voxelgridsr}, we introduce the generalizable VoxelGridSR module that serves as the core of radiance field super-resolution. We finally illustrate the training strategy for VoxelGridSR in Sec.~\ref{sec:rgbnet}. 

\subsection{Voxel-Based Distilled Feature Field}\label{sec:dff}
\begin{figure}[t!]
  \centering
  \includegraphics[width=1\columnwidth]{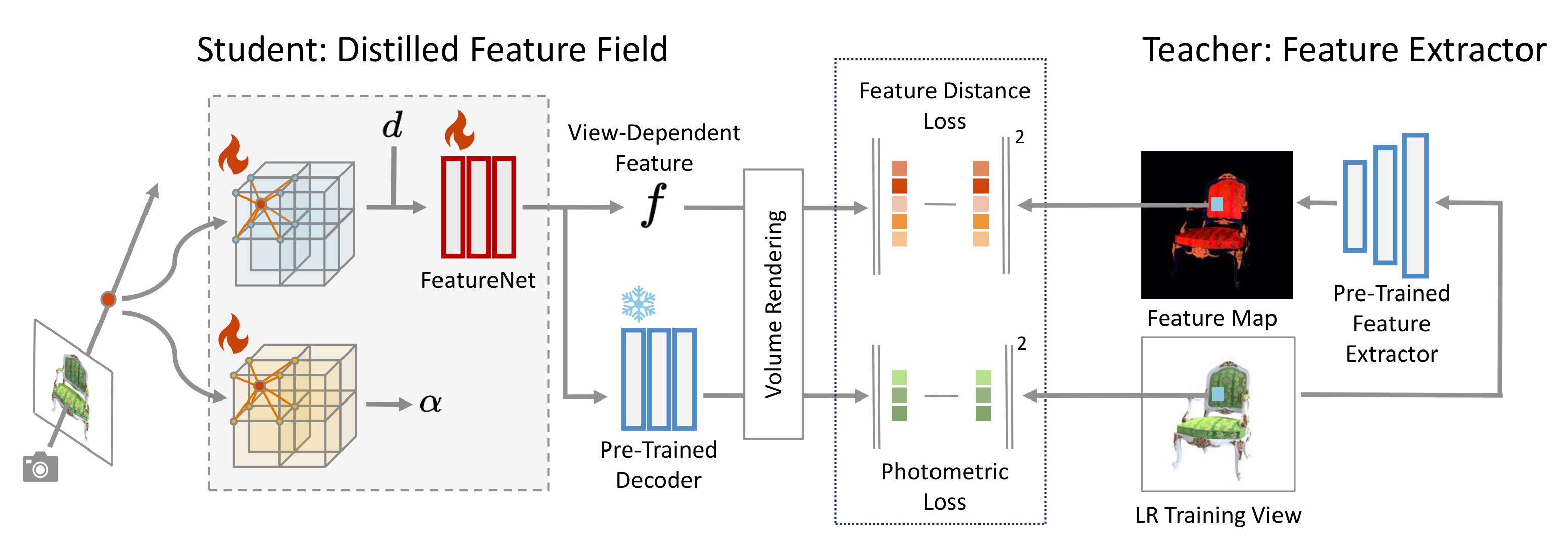}
  % \vspace{-2.5em}
  \caption{\textbf{Distilled feature field}: In a student-teacher setting, features extracted from training views are distilled into a 3D student network. The student network is trained by minimizing the difference between rendered features and features from pre-trained image feature extractor, in addition to rendered colors and ground-truth pixel colors. FeatureNet turn voxel feature into view-dependent distilled features, and a pre-trained decoder maps view-dependent features RGB color.}
  \label{fig:dff}
  % \vspace{-3em}
\end{figure}

As shown in Fig.~\ref{fig:overview}, given a query coordinate $x\in\Real^3$, density $\sigma$ and voxel feature $f$ of each of $x$'s nearest neighbors are queried from voxel grids. The feature and density voxel grids explicitly store appearance and occupancy information respectively, and FeatureNet further maps a voxel feature $f$ to a view-dependent distilled feature $f^d$ for VoxelGridSR to perform self-attention. Without special modifications, the features queried from the radiance field are nothing more than high-dimensional colors, limiting the performance of self-attention by VoxelGridSR. Applying a 3D feature extractor to the voxel feature grid is a straightforward solution, but the insufficiency of training data for a 3D feature extractor leads to poor overall performance. On the other hand, image data of large quantity as well as pre-trained models can greatly benefit tasks in 3D. Inspired by this observation, we propose distilled feature field for scene representation in ASSR-NeRF that bridges the gap between 2D and 3D data through feature distillation. 

In recent years, several works of neural radiance field have introduced feature distillation in their methods. \cite{kobayashi2022decomposing, tschernezki2022neural} add an additional branch to NeRF~\cite{mildenhall2020nerf} to learn the semantic features from DINO~\cite{caron2021emerging}, and performs open-set semantic segmentation of radiance fields with distilled semantic features on query points. \cite{kerr2023lerf} embeds multi-scale CLIP~\cite{radford2021learning} features to a radiance field and performs visual grounding in 3D space. While all the above methods distilled high-level features, we propose to distill the learning-based low-level features into our radiance field as the SR priors since they represent vast information about textures and details of scenes. Feature distillation is also a crucial to our work as it guarantees the generalizability of VoxelGridSR model. By distilling features from the same teacher extractor into the radiance fields, feature voxel grids from different scenes can have aligned same latent space, i.e., the distributions of queried features from all radiance fields remain the same. In this way, VoxelGridSR can be trained in a unified latent space shared across voxel grids from all scenes and achieve generalizability.

We first follow the autoencoder training paradigm in ~\cite{chen2021learning} to train a residual dense network (RDN)~\cite{8578360} as feature extractor $\mathrm{FE}$ and a decoder $D$. Then, we follow a student-teacher setting to distill features into radiance field, as shown in Fig.~\ref{fig:dff}. The distilled feature field is based on voxel grids, so VoxelGridSR module can directly perform super-resolution on scene representation, guaranteeing multi-view consistency of rendered novel views. Following ~\cite{sun2022direct, yu2021plenoxels}, we employ two voxel grids: a density voxel grid $V_d \in \Real^{1\times N_x \times N_y \times N_z}$ and a feature voxel grid $V_f \in \Real^{C\times N_x \times N_y \times N_z}$, to explicitly represent geometry and appearance of a 3D scene respectively, where $C$ is the dimension of feature-space. 
Given a query point $\xquery$, its density $\sigma_q$ and voxel feature $f_q'$ are queried from the voxel grids with trilinear interpolation, and FeatureNet further maps $f_q'$ and viewing direction $d$ to view-dependent features $f_q$. In addition, $D$ decodes $f_q$ to color $c$. The distilled feature field is trained by minimizing the difference between rendered features $\mathbf{\hat{F}}(r)$ and teacher's extracted features $\mathbf{F}(r)$, as well as rendered colors and ground-truth pixel colors. The total loss $L$ then becomes the sum of photometric loss $L_{photo}$ and feature distance loss $L_{feat}$:
\begin{subequations}
\begin{align}
    L &= L_{\mathrm{photo}} + \lambda L_{\mathrm{feat}}, \\
    L_{\mathrm{photo}} &= \sum_{r\in R}\| \mathbf{C}(r) - \mathbf{\hat{C}}(r)\|_2^2 ~, \ \mathbf{\hat{C}}(r) = \sum_{i=1}^{K} T_i\, \alpha_i\, D(f_i)
    \\
    L_{\mathrm{feat}} &= \sum_{r\in R}\| \mathbf{F}(r) - \mathbf{\hat{F}}(r)\|_2^2 ~, \ \mathbf{\hat{F}}(r) = \sum_{i=1}^{K} T_i\, \alpha_i\, f_i
\end{align}
\end{subequations}
where $\lambda$ is the weight of feature distance loss, and is set to 0.5 by default. Following ~\cite{kobayashi2022decomposing}, we apply stop-gradient to density when rendering $\mathbf{\hat{F}}(r)$ since $\mathbf{F}(r)$ may not be multi-view consistent.

\subsection{VoxelGridSR}\label{sec:voxelgridsr}
We detail the architecture of VoxelGridSR for refining the radiance field.
Inspired by ASSR methods~\cite{chen2021learning, cao2023ciaosr}, we design our VoxelGridSR as a local 3D implicit function that maps a 3D coordinate and its nearby voxel features to a refined feature for the later color decoding.
% VoxelGridSR performs \textit{Density-Distance-Aware Attention} to refine $f_q$, leading to finer details on rendered views.
% Inspired by \cite{shi2021pvrcnn, shi2022pvrcnn, mao2021voxel}, which introduce point-wise attention in point cloud object detection task, we propose to apply attention considering queried distilled features, densities and relative distances.
To this end, we introduce a \emph{Density-Distance-Aware Attention} for the 3D refining procedure.
Given a query point $\xquery$, we trilinear interpolate the corresponding distilled feature $f_q$ (Sec.~\ref{sec:dff}) to produce the attention query.
We gather the contextual information from the eight nearest neighbor grid point positions $\{\xith\}_{i=1}^8$, each of which comprises its distilled feature $\fith$, volume density $\sigmaith$, and the offset to the query $\sith = \xith - \xquery$.
% In addition, the geometric information are 
% we also obtain densities $\sigmaith$, features $\fith$ and relative distances of the eight nearest points $\{\xith\}_{i=1}^8$.
% Relative distances from $\xquery$ to $\xith$ can then be calculated by $s_i = x_i - x_q$.
The query, key, and value in attention operation can then be defined as: 
\begin{equation}
\begin{cases}
  \Qbf =& \mathrm{MLP}_q\left(f_q\right)\\
  % \Kbf =& \mathrm{Stack}\left( \left\{ \mathrm{MLP}_k\left([\fith; \sith; \sigmaith]\right) \right\}_{i=1}^8 \right)\\
  % \Vbf =& \mathrm{Stack}\left( \left\{ \mathrm{MLP}_v\left([\fith; \sith; \sigmaith]\right) \right\}_{i=1}^8 \right)\\
  \Kbf_i =& \mathrm{MLP}_k\left([\fith; \sith; \sigmaith]\right) \\
  \Vbf_i =& \mathrm{MLP}_v\left([\fith; \sith; \sigmaith]\right)\\
\end{cases} ~,
\end{equation}
where $\fith$, $\sith$ and $\sigmaith$ are concatenated before the MLPs, which then forms the key matrix $\Kbf\in\Real^{8 \times D_K}$ and the value $\Vbf\in\Real^{8 \times D_K}$ matrices.
Attention can then be performed by scaled dot-product attention:  
\begin{equation}
    \frefine_q = \softmax\left( \frac{\mathbf{Q} \cdot \mathbf{K}\trans}{\sqrt{D_K}} \right)\mathbf{V}
\end{equation}
where $D_K$ is the dimension of voxel feature.
The \emph{Density-Distance-Aware Attention} allow the VoxelGridSR to take the feature relevancy and the local spatial relationship into consideration. Density information is also beneficial because it helps differentiate interfaces, \eg, objects and air.
The distilled feature from \cref{sec:dff} enables VoxelGridSR to utilize the SR prior to enhance the textures and details of the scene.
% We would like to point out that without the design of feature distillation, VoxelGridSR only performs attention on voxel features lacking information of textures and details of objects, leading to poor generalizability and performance.
Additionally, we also refine the geometry by aggregating the grid point densities with the attention weights:
\begin{equation}
    \sigma_q = \softmax\left( \frac{\mathbf{Q} \cdot \mathbf{K}\trans}{\sqrt{D_K}} \right) \cdot [\sigma_1, \cdots, \sigma_8]\trans
\end{equation}

\subsection{Multi-Scene Training for VoxelGridSR}\label{sec:rgbnet}
\begin{figure*}[t]
	\centering
	\includegraphics[width=1\columnwidth]{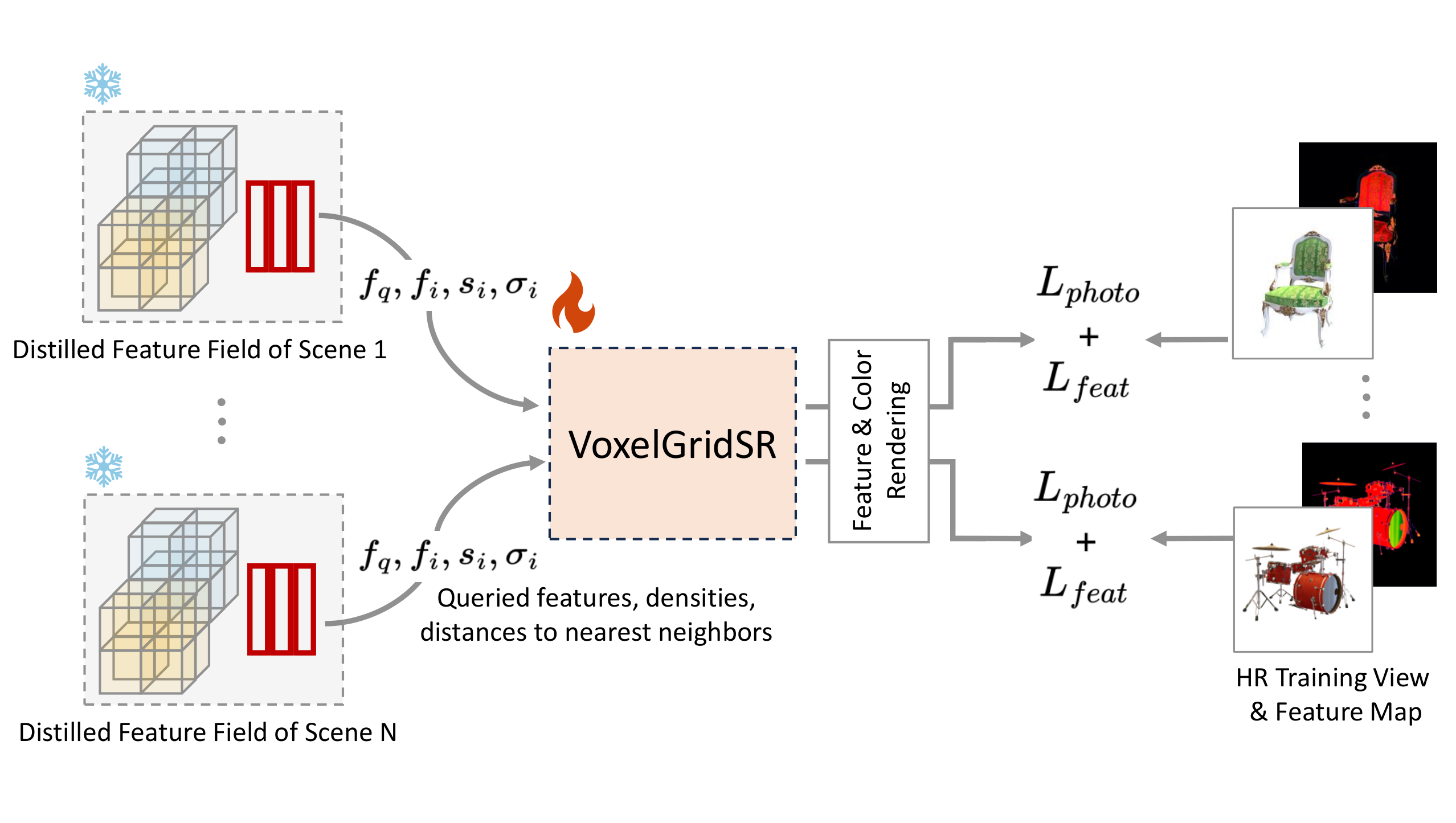}
 % \vspace{-10mm}
    \caption{\textbf{Multi-scene training for VoxelGridSR}: We train the generalizable VoxelGridSR module with $N$ distilled feature fields that are reconstructed from LR training views as input and take HR training views as well as feature maps as ground-truth. In every iteration of training, a feature field of scene $i\in N$ is randomly selected. Then, VoxelGridSR maps $f_q$ to $\frefine_q$ and $c$, and is updated by $L_{photo}$ and $L_{feat}$.
    }\label{fig:training}
    % \vspace{-5mm}
\end{figure*}
% In the previous sections, we introduce the components of ASSR-NeRF: a distilled feature field and a VoxelGridSR module. We want to highlight that the VoxelGridSR module is generalizable, meaning that whoever trained a distilled feature field with any set of LR training views can directly apply the pre-trained VoxelGridSR to the feature field and perform HRNVS. 
Training VoxelGridSR on a single scene is less useful.
As the distilled 3D feature is aligned to the SR latent space of the 2D teacher, we train VoxelGridSR on multiple scenes for generalizability.
Once trained, the VoxelGridSR module can serve as an off-the-shelf enhancer to any distilled feature field under the same latent space. Fig.~\ref{fig:training} depicts the multi-scene training procedure of VoxelGridSR. We first pre-trained $N$ scenes with LR views and distilled feature fields.
% During each iteration of training, a feature field of scene $i \in N$ is randomly selected, and VoxelGridSR refines feature $f_q$ of every queried point $x_q$ to $\frefine_q$ through \textit{Density-Distance-Aware Attention}.
Subsequently, during each iteration of the cross-scene training, a feature field of scene $i \in N$ is randomly selected and refined by VoxelGridSR through \textit{Density-Distance-Aware Attention}.
We then can minimize the photometric loss $L_{photo}$ between the rendered SR view and the ground-truth HR view.
Feature matching loss $L_{feat}$ is also applied to ensure the latent space to remain consistent.

\section{Experiments}
\label{sec:experiments}

\subsection{Implementation Details}\label{sec:implementation}
We impelment ASSR-NeRF with PyTorch~\cite{paszke2019pytorch}. To sample all the densities, features and relative distances of nearest neighbors efficiently given a 3D position, we design custom CUDA extensions. 
We set an expected number of voxels $256^3$ for both density and feature voxel grids. FeatureNet in our distilled feature field is based on RDN~\cite{8578360}, and the pre-trained decoder consists of 5 MLP layers with dimensions of $256$. FeatureNet and decoder are trained together in an autoencoder paradigm on DIV2K~\cite{8014883} dataset. To train VoxelGridSR, we take the BlendedMVS~\cite{yao2020blendedmvs} dataset at resolution $768\times576$ as HR ground-truth training views and downsample with a factor $\times4$ to generate corresponding LR training views. Distilled feature field of every scene is trained for $20k$ iterations and a batchsize $4096$. We first reconstructed distilled feature fields of $40$ scenes from BlendedMVS. After reconstructing distilled feature fields of $40$ scenes, we train the VoxelGridSR module with the distilled feature fields for $240K$ and a batchsize $2048$.
 
\subsection{Comparisons and Discussions}
\label{sec:comparison}
We compare ASSR-NeRF and other state-of-the-art NeRF-based NVS methods~\cite{sun2022direct,barron2023zipnerf} with a few different settings. Among them, DVGO~\cite{sun2022direct} serves as a baseline since its architecture, consisting voxel grids and shallow MLP layers, is similar to us. Zip-NeRF~\cite{barron2023zipnerf} is optimized for anti-aliasing so that their rendered novel views remain clear with large difference in training and testing resolutions. We also compare our method with image SR methods~\cite{Liang2021SwinIRIR, wang2023exploiting} and a radiance field SR method~\cite{10205145}.

\subsubsection{NeRF-based NVS methods}\label{sec:multi_compare}

\begin{table*}[hbt!]
	\centering
	\caption{Quantitative results on Synthetic-NeRF ~\cite{mildenhall2020nerf} and BlendedMVS ~\cite{yao2020blendedmvs}: We compare ASSR-NeRF with other NeRF-based NVS methods. All methods are trained with LR training views of a downsampling factor x4, and render at three different resolutions. ASSR-NeRF achieves the best performance in all resolutions and datasets.}
	\label{tab:blender}
 \resizebox{\textwidth}{!}{%
	\begin{tabular}{|c||c c c| c c c|c c c|}
		\hline
    \ &\multicolumn{3}{|c}{x1.6}
        &\multicolumn{3}{|c}{x2} &\multicolumn{3}{|c|}{x4}\\
		\cline{2-10}
		  $\textit{Synthetic-NeRF}$ &  PSNR $\uparrow$ & SSIM $\uparrow$  & LPIPS $\downarrow$ & PSNR $\uparrow$ & SSIM $\uparrow$   & LPIPS $\downarrow$ & PSNR $\uparrow$ & SSIM $\uparrow$   & LPIPS $\downarrow$  \\
		\hline\hline
  %           TensoRF \cite{chen2022tensorf} & 24.65 & 0.771 & 0.268 & 22.78 & 0.658 & 0.417 & \textbf{22.29} & \textbf{0.632} & 0.467 & \textbf{22.29} & \textbf{0.656} & \textbf{0.489}\\
		% \hline
  %           Instant-ngp \cite{M_ller_2022} & \textbf{0.909} & \textbf{0.906} & 0.860 & 0.875 & 0.840 & 0.841 & dsf & adf & 0.840 & 0.841 & dsf & adf\\
		% \hline
            DVGO \cite{sun2022direct} & 29.89 & 0.945 & 0.061 & 28.80 & 0.933 & 0.077 & 27.33 & 0.910 & 0.115 \\
		\hline
            TensoRF \cite{chen2022tensorf} & \textbf{31.36} & 0.950 & 0.060 & 29.96 & 0.947 & 0.078 & 28.07 & 0.916 & 0.118\\
		\hline
            Instant-ngp \cite{M_ller_2022}
            & 28.55 & 0.933 & 0.095 & 28.23 & 0.926 & 0.101 & 27.26 & 0.902 & 0.128\\
            \hline
            Zip-NeRF \cite{barron2023zipnerf}
            & 30.95 & \textbf{0.962} & \textbf{0.041} & 29.56 & 0.951 & 0.057 & 27.73 & 0.923 & 0.102\\
		\hline\hline
            ASSR-NeRF (ours)
            & 31.09 & 0.961 & 0.048 & \textbf{30.57} & \textbf{0.954} & \textbf{0.057} & \textbf{29.02} & \textbf{0.932} & \textbf{0.093}\\
		\hline
		
	\end{tabular}
}

\vspace*{0.2 cm}

 \resizebox{\textwidth}{!}{%
	\begin{tabular}{|c||c c c| c c c|c c c|}
		\hline
    \ &\multicolumn{3}{|c}{x2}
        &\multicolumn{3}{|c}{x2.5} &\multicolumn{3}{|c|}{x4}\\
		\cline{2-10}
		  $\textit{BlendedMVS}$ &  PSNR $\uparrow$ & SSIM $\uparrow$  & LPIPS $\downarrow$ & PSNR $\uparrow$ & SSIM $\uparrow$   & LPIPS $\downarrow$ & PSNR $\uparrow$ & SSIM $\uparrow$   & LPIPS $\downarrow$  \\
		\hline\hline
  %           TensoRF \cite{chen2022tensorf} & 24.65 & 0.771 & 0.268 & 22.78 & 0.658 & 0.417 & \textbf{22.29} & \textbf{0.632} & 0.467 & \textbf{22.29} & \textbf{0.656} & \textbf{0.489}\\
		% \hline
  %           Instant-ngp \cite{M_ller_2022} & \textbf{0.909} & \textbf{0.906} & 0.860 & 0.875 & 0.840 & 0.841 & dsf & adf & 0.840 & 0.841 & dsf & adf\\
		% \hline
            DVGO \cite{sun2022direct} 
            & 26.88 & 0.909 & 0.111 & 26.96 & 0.902 & 0.124 & 24.74 & 0.845 & 0.187\\
		\hline
            TensoRF \cite{chen2022tensorf} & 27.72 & 0.921 & 0.114 & 27.78 & 0.912 & 0.131 & 25.35 & 0.852 & 0.182\\
		\hline
            Instant-ngp \cite{M_ller_2022}
            & 26.63 & 0.894 & 0.150 & 26.64 & 0.882 & 0.166 & 25.14 & 0.835 & 0.188\\
            \hline
            Zip-NeRF \cite{barron2023zipnerf}
            & 28.48 & 0.929 & 0.148 & 28.41 & 0.918 & 0.173 & 25.97 & 0.854 & 0.237\\
		\hline\hline
            ASSR-NeRF (ours)
            & \textbf{28.52} & \textbf{0.931} & \textbf{0.080} & \textbf{28.56} & \textbf{0.926} & \textbf{0.093} & \textbf{26.38} & \textbf{0.873} & \textbf{0.148}\\
		\hline
		
	\end{tabular}
}               
\end{table*}

\begin{table*}[hbt!]
	\centering
	\caption{Quantitative results on BlendedMVS~\cite{yao2020blendedmvs}: We compare our method with hyrbrid approaches where LR rendered views are super-resolved by image SR methods with factor x4 and a NeRF SR method that shares the same setting with us.}
	\label{tab:blendedmvs}
 \resizebox{8cm}{!}{%
	\begin{tabular}{|c||c c c|}
            \hline
		  $\textit{}$ &  PSNR $\uparrow$ & SSIM $\uparrow$  & LPIPS $\downarrow$   \\
		\hline\hline
  %           TensoRF \cite{chen2022tensorf} & 24.65 & 0.771 & 0.268 & 22.78 & 0.658 & 0.417 & \textbf{22.29} & \textbf{0.632} & 0.467 & \textbf{22.29} & \textbf{0.656} & \textbf{0.489}\\
		% \hline
  %           Instant-ngp \cite{M_ller_2022} & \textbf{0.909} & \textbf{0.906} & 0.860 & 0.875 & 0.840 & 0.841 & dsf & adf & 0.840 & 0.841 & dsf & adf\\
		% \hline
            Zip-NeRF \cite{barron2023zipnerf} + SwinIR \cite{Liang2021SwinIRIR} 
            & 26.21 & 0.866 & 0.159\\
		\hline
            Zip-NeRF \cite{barron2023zipnerf} + StableSR \cite{wang2023exploiting}
            & 24.56 & 0.839 & 0.169 \\
		\hline
            CROP \cite{10205145}
            & 26.25 & \textbf{0.879} & \textbf{0.145}    \\
		\hline\hline
            ASSR-NeRF (ours)
            & \textbf{26.38} & 0.873 & 0.148   \\
		\hline
		
	\end{tabular}
}
\end{table*}
\begin{figure*}[h!]
	\centering
	\includegraphics[width=1\columnwidth]{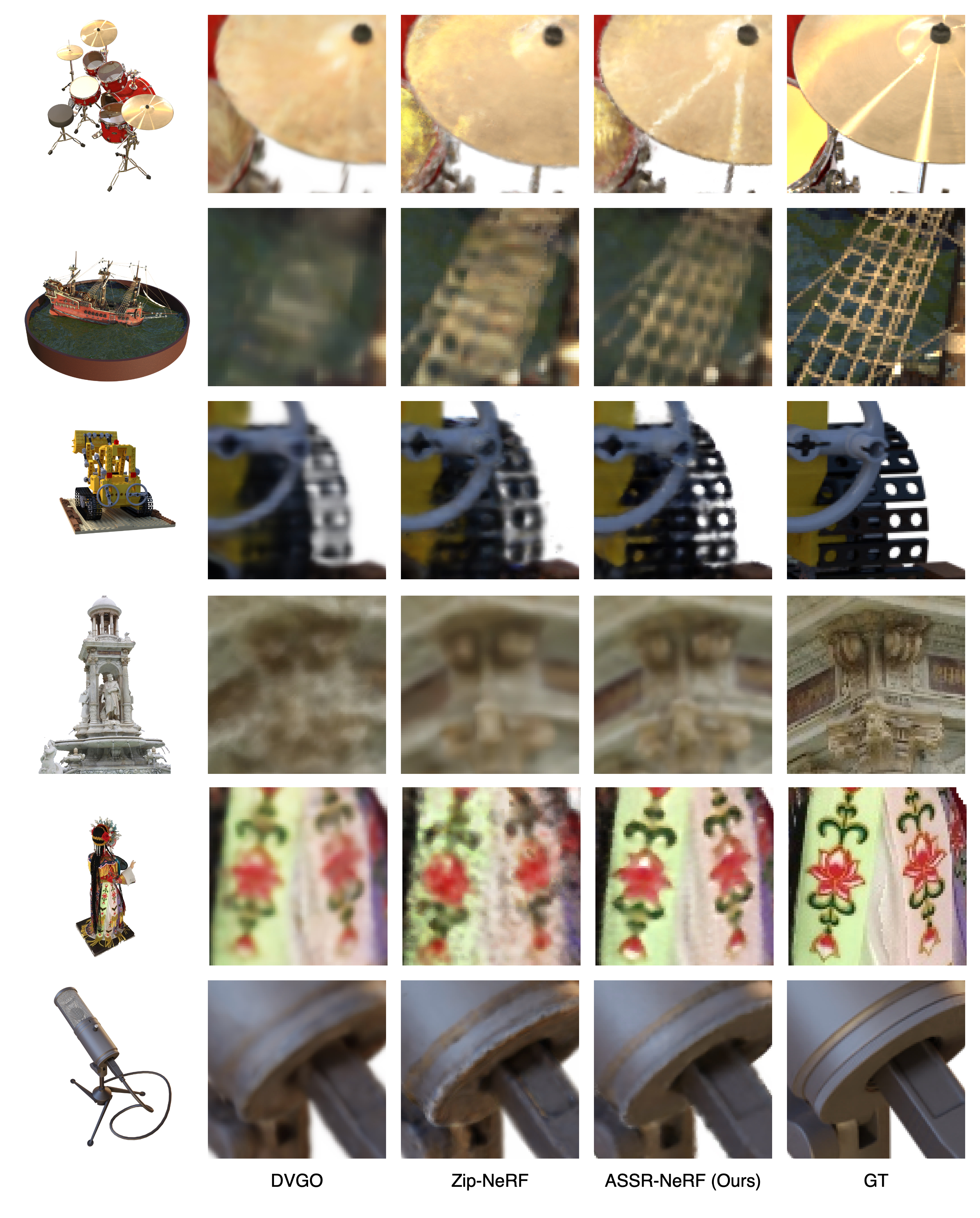}
 % \vspace{-2mm}
    \caption{Qualitative results on Synthetic-NeRF~\cite{zhang2020nerf} and BlendedMVS~\cite{yao2020blendedmvs}: All other baselines are first reconstructed from LR training views, then perform HRNVS. For ASSR-NeRF, pre-trained VoxelGridSR model is applied to achieve SRNVS. The results show that ASSR-NeRF generates cleaner edges as well as richer details than other baselines. 
    }
    % \vspace{-3mm}
	\label{fig:quality}
\end{figure*}
In this experiment, we compare ASSR-NeRF with state-of-the-art NeRF-based methods for high-resolution novel view synthesis (HRNVS). For every scene, we train a distilled feature field with LR training views, and directly apply pre-trained VoxelGridSR on the distilled feature field to perform super-resolution novel view synthesis (SRNVS). We also train other methods with LR training views and perform HRNVS. We compare all the methods on two datasets, Synthetic-NeRF~\cite{mildenhall2020nerf} and BlendedMVS~\cite{yao2020blendedmvs} with three different scales. Note that there is no overlapping between scenes used to train VoxelGridSR and scenes used for testing in all experiments. In Tab.~\ref{tab:blender}, we show the result in PSNR, SSIM and LPIPS. We can see that ASSR-NeRF surpasses all the other methods at every scale. Although Zip-NeRF gains some advantage when the scale is low, its performance declines greatly when the scale increases, revealing common NeRF-based methods' shortcomings. This result matches our initial observation that NVS methods trained on LR training views have non-ideal performance when rendering HR novel views. We also show the qualitative results in Fig.~\ref{fig:quality}. While having cleaner rendered views, ASSR-NeRF can also generate finer details as well as textures. For example, ASSR-NeRF generates cleaner patterns on figure's clothes and sharper edge of microphone.

\subsubsection{Super-resolution methods}
\label{sec:single_compare}
In this experiment, we compare our method with image SR methods as well as radiance field SR methods. Following the training setting from Sec.~\ref{sec:multi_compare}, we train Zip-NeRF~\cite{barron2023zipnerf} with LR training views. Then, we render LR novel view at the same scale, and super-resolve the novel views with state-of-the-art image SR methods~\cite{Liang2021SwinIRIR,wang2023exploiting}. SwinIR~\cite{Liang2021SwinIRIR} integrates Swin Transformer~\cite{liu2021swin} into its architecture for better feature extraction on input images. StableSR~\cite{wang2023exploiting} utilized pre-trained diffusion models to generate finer details and textures on SR results. Besides using image SR on rendered novel views, we compare with a radiance field SR method~\cite{10205145} that shares a similar setting with us. CROP~\cite{10205145} first super-resolves LR training views with pre-trained image SR methods, and uses the super-resolved views to train a NeRF-based NVS model. Unlike other radiance field SR methods~\cite{Wang_2022, 10205402} that requires an HR reference image of the same scene, CROP only needs LR training views when testing. Our method differs from CROP that our VoxelGridSR module is optimized at arbitrary scale while CROP is trained at a fixed upscaling factor. All methods perform HRNVS with a upscaling factor $\times4$. In Tab.~\ref{tab:blendedmvs}, we show the quantatitive results on BlendedMVS~\cite{yao2020blendedmvs}. While reaching comparable SSIM and LPIPS scores as CROP, our method outperforms in terms of PSNR.

\subsubsection{Multi-View Consistency}
\label{sec:multi-view}
\begin{figure*}[t]
	\centering
	\includegraphics[width=1\columnwidth]{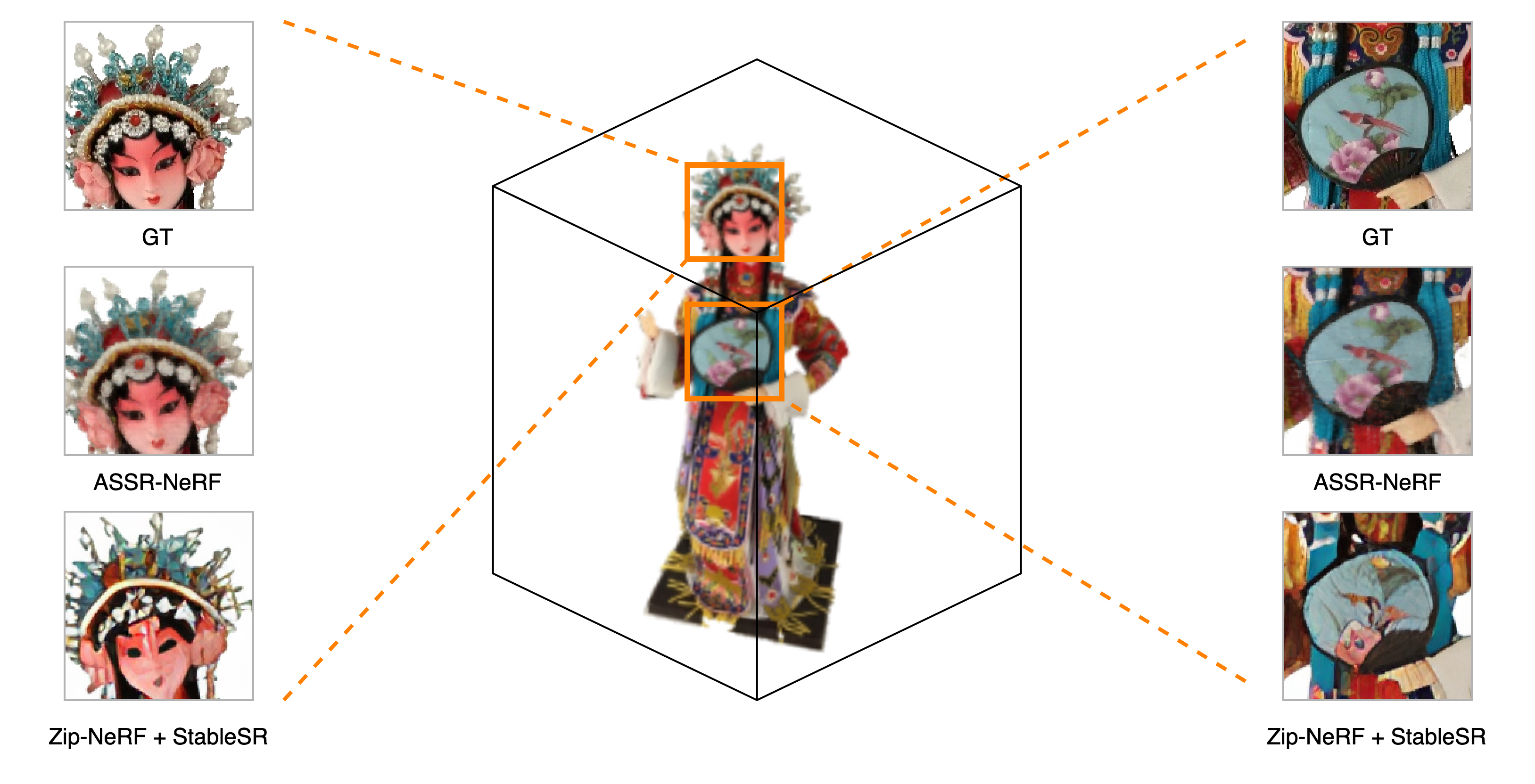}
 % \vspace{-2mm}
    \caption{Qualitative results of view consistency. Applying image SR method, e.g., StableSR~\cite{wang2023exploiting} to rendered LR novel views from Zip-NeRF ~\cite{barron2023zipnerf} leads to sharp but distorted results, and loses multi-view consistency. On the other hand, our method generates clean HR novel views with consistent details. We encourage readers to view our supplementary videos where our method achieves better multi-view consistency than other baselines.
    }
    % \vspace{-3mm}
	\label{fig:consistency}
\end{figure*}
In this section, we discuss the multi-view consistency issue. In Sec.~\ref{sec:single_compare}, we conduct the experiments of super-resolving rendered LR novel view with image SR methods. Although this approach can also generate cleaner HR novel views with finer details, multi-view inconsistency remains as a serious problem. As shown in Fig.~\ref{fig:consistency}, super-resolving rendered LR novel views with image SR methods lead to distorted geometry and textures across different views. On the other hand, our method generate consistent views from different camera poses. This advantage of our method is contributed from the design of ASSR-NeRF's SR module. Instead of applying SR on 2D feature maps, VoxelGridSR directly applies SR on 3D volume, i.e., the distilled feature fields, guaranteeing consistency of geometry and appearance across every viewing direction.

\subsection{Ablation Studies}
\label{sec:ablation}
We provide ablation studies to analyze our proposed method in this section. We analyze the design of VoxelGridSR module to verify the effectiveness of \textit{Density-Distance-Aware Attention}. 

\begin{table}[hbt!]
	\centering
	\caption{Ablation study on VoxelGridSR module. We verify the effectiveness of density-aware and distance-aware attention on voxel grids. While considering both denstiy and relative distance of nearest neighbors improves performance, we highlight that density information is more crucial to SR in 3D space.}
	\label{tab:ablation}
 % \resizebox{10cm}{!}{%
	\begin{tabular}{|c||c c c| c c c|}
		\hline
		  Model &  Feature-aware  & Density-aware  & Distance-aware  & PSNR $\uparrow$ & SSIM $\uparrow$   & LPIPS $\downarrow$   \\
		\hline\hline
  %           TensoRF \cite{chen2022tensorf} & 24.65 & 0.771 & 0.268 & 22.78 & 0.658 & 0.417 & \textbf{22.29} & \textbf{0.632} & 0.467 & \textbf{22.29} & \textbf{0.656} & \textbf{0.489}\\
		% \hline
  %           Instant-ngp \cite{M_ller_2022} & \textbf{0.909} & \textbf{0.906} & 0.860 & 0.875 & 0.840 & 0.841 & dsf & adf & 0.840 & 0.841 & dsf & adf\\
		% \hline
            A
            & \checkmark & \checkmark & \checkmark & 26.38 & 0.873 & 0.148   \\
		\hline
            B
            & \checkmark & \checkmark &  & 26.20 & 0.873 & 0.157  \\
		\hline
            C
            & \checkmark &  & \checkmark & 26.14 & 0.872 & 0.147  \\
		\hline
		
	\end{tabular}
% }
\end{table}
\subsubsection{Analysis of VoxelGridSR}
\label{sec:voxelgridsr_ablation}
We follow the same training setting described in Sec.~\ref{sec:implementation} and train different variants of VoxelGridSR. Tab.~\ref{tab:ablation} shows the variants' performance of SRNVS on BlendedMVS dataset with an upscaling factor $\times4$. \textit{Density-aware} means whether to consider densities of nearest neighbors when performing self-attention, and \textit{Distance-aware} indicates whether the relative distance to each nearest neighbor is considered. The metrics show that both density and relative distance can benefit the performance of self-attention. Interestingly, we find that \textit{Distance-aware} brings more improvements for 3D SR. This is reasonable, since density contains information about the local geometry and SR in 3D space can thus lead to sharper edges of objects in rendered views.

% \subsubsection{Analysis of distilled feature field}
% \label{sec:dff_ablation}
% We also analyze our proposed distilled feature field, which is radiance field distilled with image feature maps from teacher feature extractor. Described in Sec.~\ref{sec:dff}, the feature distillation technique enables \textit{Density-Distance-Aware Attention} to operate on meaningful features from distilled feature field, leading to better performance. In addition to this benefit, feature distillation is also indispensable to train the generalizable VoxelGridSR module. In voxel-based radiance field, queried features from feature voxel grid are turned into view-dependent color $c$ by a decoder. Without feature distillation, the latent distribution in feature voxel grid of every scene will be different, since each radiance field is optimized independently with its own decoder. With feature distillation, all voxel grids across scenes will share the same latent distribution, i.e., the latent distribution of the teacher feature extractor. As a result, VoxelGridSR can be trained in a unified latent space shared across voxel grid from every scene and finally achieve generalizability. Fig.xxx shows the importance of feature distillation that VoxelGridSR module trained on normal voxel-based radiance field can only generate unsatisfied results. 

\subsection{Discussions and limitations}\label{sec:discussions}
The experiments show that our framework achieves competitive performance in SRNVS. However, one of limitations is the increased rendering time since VoxelGridSR performs self-attention on every sampled point. Reducing rendering time while keeping the same quality will be our future research direction. We also notice that currently there isn't an effective benchmark to assess multi-view consistency, and we can only compare our methods with others through frame-wise metrics such as PSNR and qualitative presentations. We think video quality assessment (VQA) might be an interesting research direction that can serve as an assessment metric for multi-view consistency.

\section{Conclusions}
\label{sec:conclusion}

In this work, we propose ASSR-NeRF, a novel framework for radiance field super-resolution. ASSR-NeRF consists of a distilled feature field for scene representation and a generalizable VoxelGridSR module for raidance field SR. Once a distilled feature field is reconstructed from any set of LR training views, a pre-trained generalizable VoxelGridSR module can be directly applied for super-resolution novel view synthesis (SRNVS). Our approach can greatly benefit real-world applications. For example, low-resolution training views captured by cheap devices can be efficiently utilized for high-quality novel view synthesis, reducing cost and time required by views capturing. Experiments on various benchmarks as well as qualitative comparisons show that our framework is strongly effective in improving rendering quality.

% \clearpage  % TODO REVIEW/FINAL: This \clearpage needs to be removed from both review and camera-ready versions.

\bibliographystyle{splncs04}
\bibliography{main}

\title{Supplementary Material}
\author{}
\institute{}

\titlerunning{Supplementary Material for ASSR-NeRF}

\maketitle
\vspace{-8mm}
\section{Additional Implementation Details}
\label{sec:add_implementation}

\subsection{Feature extractor and pre-trained decoder}

As described in Sec.4 of the main paper, we use a pre-trained feature extractor $\mathrm{FE}$ to provide SR priors to be distilled into radiance field and a pre-trained decoder $D$ to map voxel features to distilled features. We follow the training procedure from \cite{chen2021learning} to train $\mathrm{FE}$ and $D$ together. Under an autoencoder paradigm, an input image $I$ of shape $\mathrm{H}\times \mathrm{W}\times 3$ is first encoded into a pixel-aligned feature map $F$ of shape $\mathrm{H}\times \mathrm{W}\times \mathrm{C}$ by $\mathrm{FE}$, where $\mathrm{H}$ and $\mathrm{W}$ is the height and width of the input image and $\mathrm{C}$ is the dimension of the feature. The decoder $D$ then maps pixel-aligned feature to RGB value. The models can then be trained to minimize the L1 loss between the decoder output and the input image $I$. We choose RDN~\cite{8578360} as $\mathrm{FE}$ for its good trade-off of training efficiency and performance compared to other architectures and a decoder $D$ of 5 MLP layers with ReLU activation function. 

\subsection{Preprocessing of dataset}

\begin{figure}[h!]
	\centering
	\includegraphics[width=1\columnwidth]{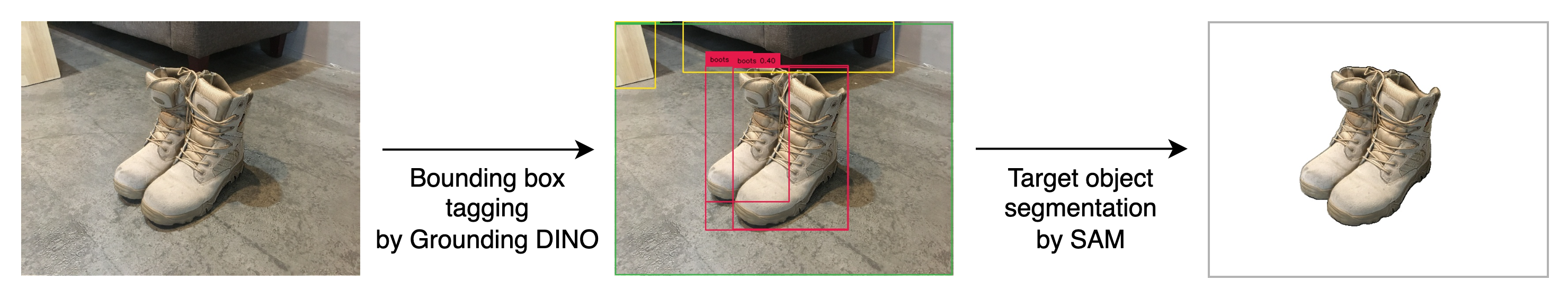}
 % \vspace{-10mm}
    \caption{\textbf{Pipeline for dataset preprocessing}: Given a raw training view, we first use Gounding DINO~\cite{liu2023grounding} to locate the target obejct, then utilize a segment anything model (SAM)~\cite{kirillov2023segment} to segment and generate training view with object mask.
    }\label{fig:preprocessing}
    % \vspace{-5mm}
\end{figure}

We train the generalizable VoxelGridSR model for all experiments with BlendedMVS~\cite{yao2020blendedmvs} dataset, and test our method on a subset of 5 scenes, following all previous works. Before training VoxelGridSR model, we first reconstructed 40 radiance fields of scenes from BlendedMVS. We found that BlendedMVS mostly contain scenes of complex objects as well as complicated backgrounds, which may affect the quality of the reconstructed radiance field. To ensure that VoxelGridSR is trained with well-reconstructed radiance fields, we design a preprocessing pipeline to refine BlendedMVS data. Given a raw training view of scene $s$, we first use Grounding DINO~\cite{liu2023grounding} to tag the bounding box of target object. Then, a segment anything model (SAM)~\cite{kirillov2023segment} is used to segment the object and generate training view with object mask. The processed images and camera poses are then used to reconstruct radiance fields for training VoxelGridSR. Pre-trained Grounding DINO and SAM are directly utilized for the procedure.

\section{Multi-View Consistency}
\label{sec:multiview}

\begin{figure}[h!]
	\centering
	\includegraphics[width=0.8\columnwidth]{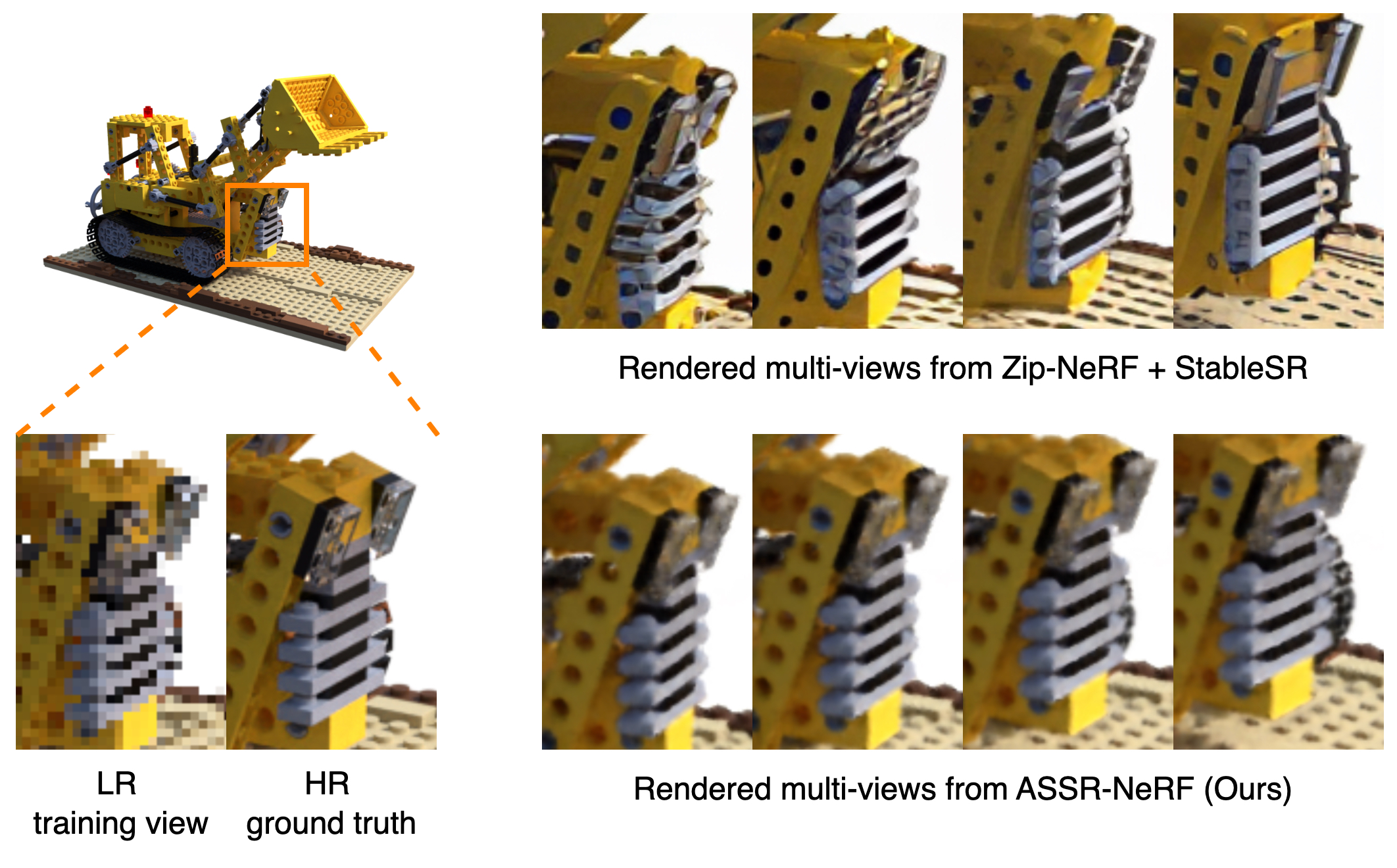}
 \vspace{-2mm}
    \caption{\textbf{Comparison of multi-view consistency}: Super-resolving LR novel views from Zip-NeRF ~\cite{barron2023zipnerf} by StableSR ~\cite{wang2023exploiting} leads to serious inconsistency across views from different camera poses. ASSR-NeRF can render HR novel views of consistent geometry and appearance. We encourage readers to visit our video showing the consistency issue at \url{https://drive.google.com/file/d/1h8WjmN7r1R79Cd4Q-dRLgbhToMZNR3pz/view}.
    }\label{fig:consistency_compare_supp}
    % \vspace{-5mm}
\end{figure}

We provide additional qualitative comparisons about multi-view consistency in Fig.~\ref{fig:consistency_compare_supp}. While super-resolving rendered LR novel views from Zip-NeRF ~\cite{barron2023zipnerf} leads to sharper results, it results in multi-view inconsistency, i.e., geometry and appearance across views from adjacent camera poses varies a lot. On the other hand, our method performs super-resolution novel view synthesis (SRNVS) with great consistency.
\section{Analysis of Feature Distillation}
\label{sec:analysis_feature}

\begin{figure}[h!]
	\centering
	\includegraphics[width=0.8\columnwidth]{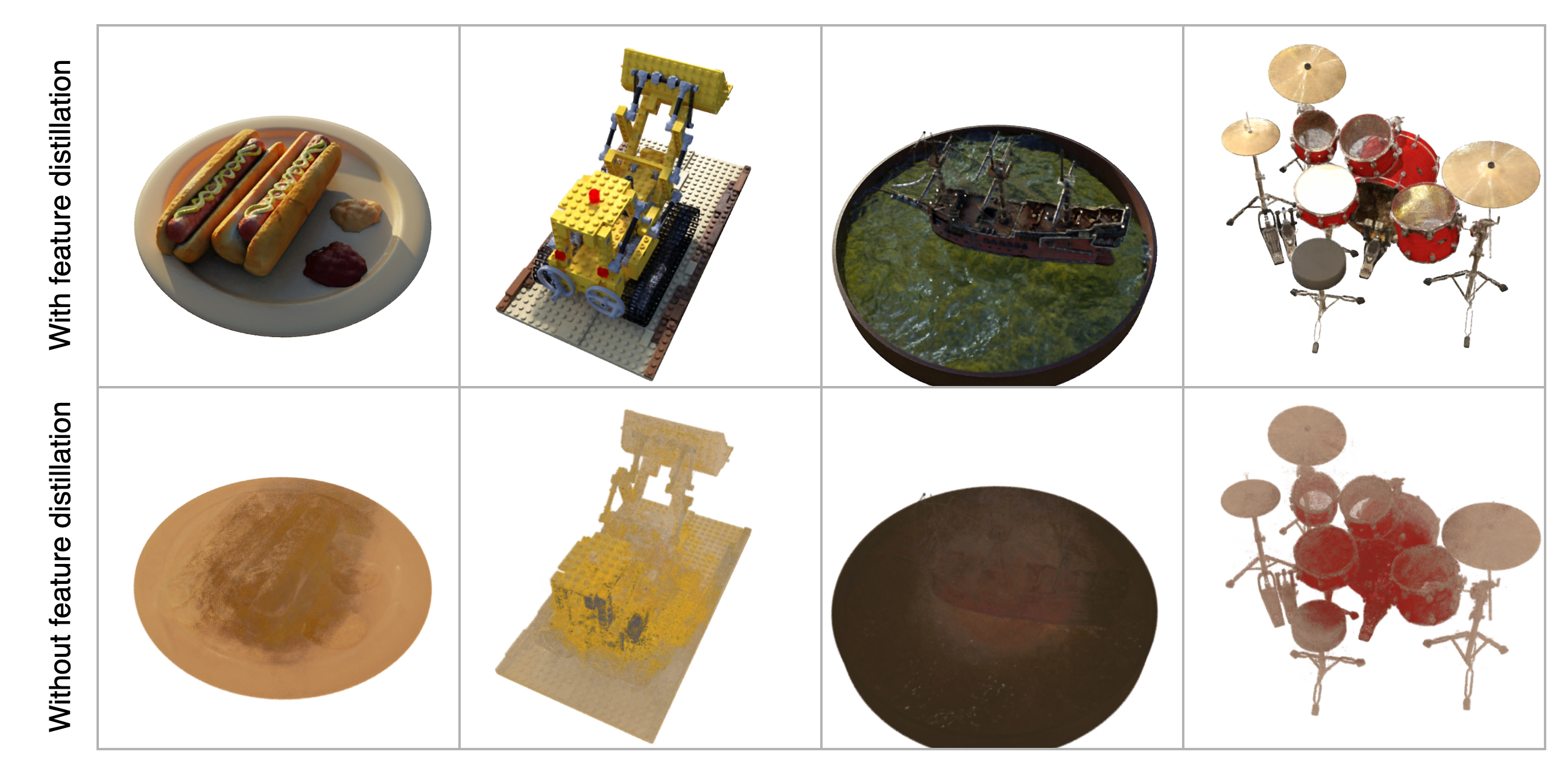}
 % \vspace{-10mm}
    \caption{\textbf{Effectiveness of training VoxelGridSR with distilled feature fields}: Shown in the upper row, VoxelGridSR model trained with distilled feature fields can achieve generalizability and perform SR on radiance fields of unseen scenes. Trained with radiance fields without feature distillation, VoxelGridSR fails to super-resolve and leads to corrupted novel views with incorrect colors, as shown in the lower row.
    }\label{fig:fd_compare}
    % \vspace{-5mm}
\end{figure}

In Sec.4.2. of the main paper, we mention that feature voxel grids from different scenes can have aligned same latent space by distilling features from the same teacher extractor into the radiance fields, VoxelGridSR can thus be trained in a unified latent space shared across voxel grids from all scenes and achieve generalizability. Fig.~\ref{fig:fd_compare} shows the importance of distilled feature fields to training a generalizable VoxelGridSR. Trained with distilled feature fields, VoxelGridSR can reach generalizability and perform SR successfully on unseen scenes, as shown in the upper row. Without feature distillation, VoxelGridSR is trained with voxel-based radiance fields of diverse voxel feature distribution, and eventually fails to gain SR ability, as shown in the lower row of Fig.~\ref{fig:fd_compare}.
\section{Additional Results}
\label{sec:additional_results}

\subsection{Bounded scenes}
\begin{figure*}[h!]
	\centering
	\includegraphics[width=1\columnwidth]{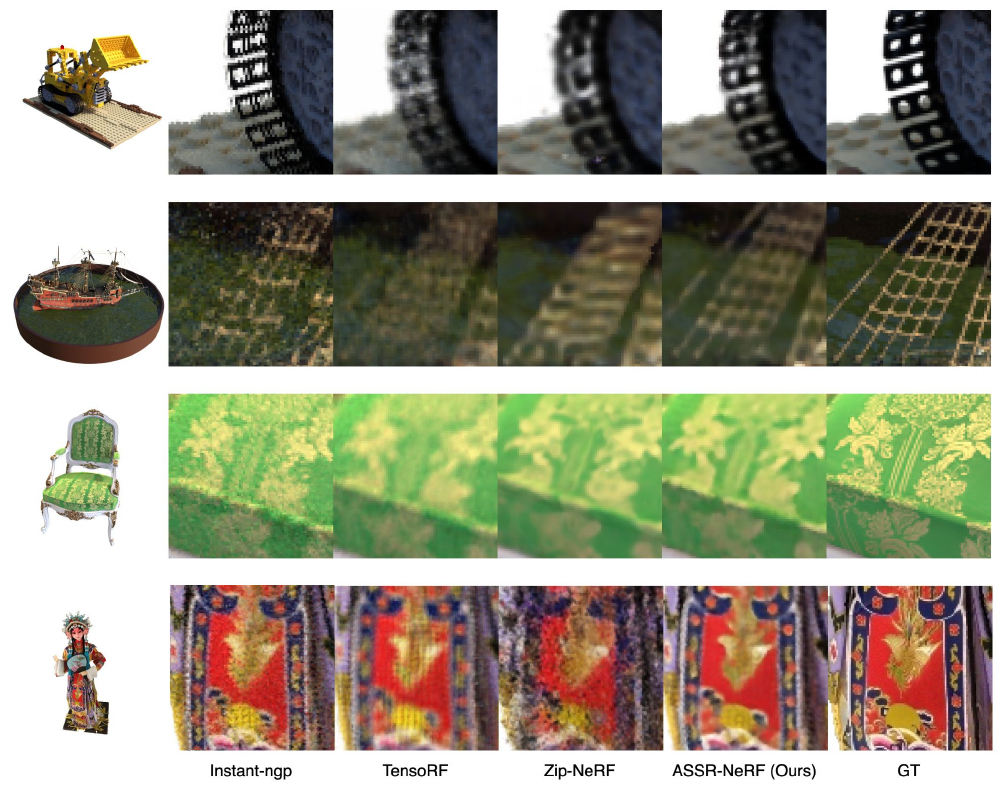}
 % \vspace{-2mm}
    \caption{Additional qualitative results: All other baselines are first reconstructed from LR training views, then perform HRNVS. For ASSR-NeRF, pre-trained VoxelGridSR model is applied to achieve SRNVS. 
    }
    % \vspace{-3mm}
	\label{fig:quality}
\end{figure*}

In Sec.5.2. of the main paper, we provide qualitative comparison with DVGO~\cite{sun2022direct} and Zip-NeRF~\cite{barron2023zipnerf}. Here we provide additional qualitative results from more models, including TensoRF~\cite{chen2022tensorf} and Instant-ngp~\cite{M_ller_2022}, on Synthetic-NeRF~\cite{mildenhall2020nerf} and BlendedMVS~\cite{yao2020blendedmvs}. Fig.~\ref{fig:quality} shows that our method can render high-resolution novel views with richer and cleaner details. 

\subsection{Forward-facing scenes}

\begin{figure}[h!]
\Huge
\centering
\resizebox{0.6\linewidth}{!}{%

\begin{tabular}{cccc}

 % {\fontsize{30}{36}\selectfont Ours (Original)}&
 % {\fontsize{30}{36}\selectfont Ours (Modified)}&
 % {\fontsize{30}{36}\selectfont GT} \\
 \rotatebox{90}{\hspace{2cm}\fontsize{50}{200}\selectfont Zip-NeRF}&
\includegraphics[]{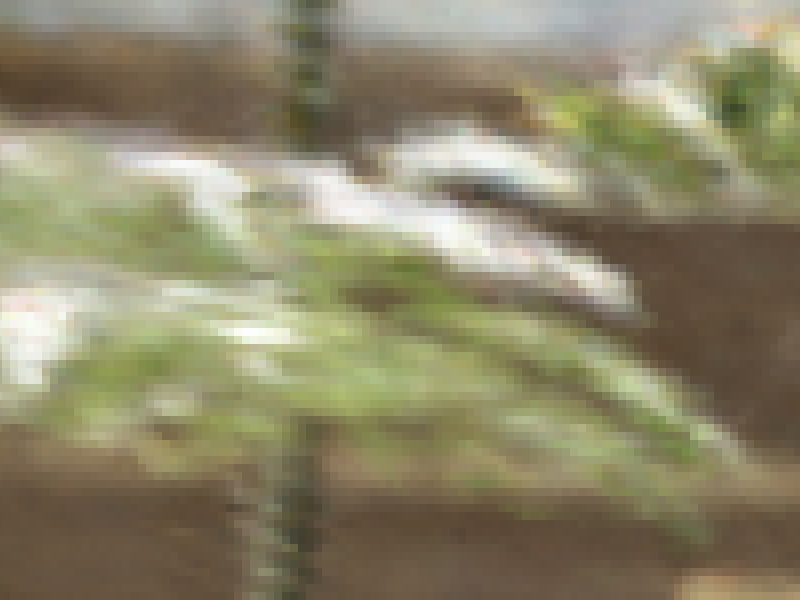}  &
\includegraphics[]{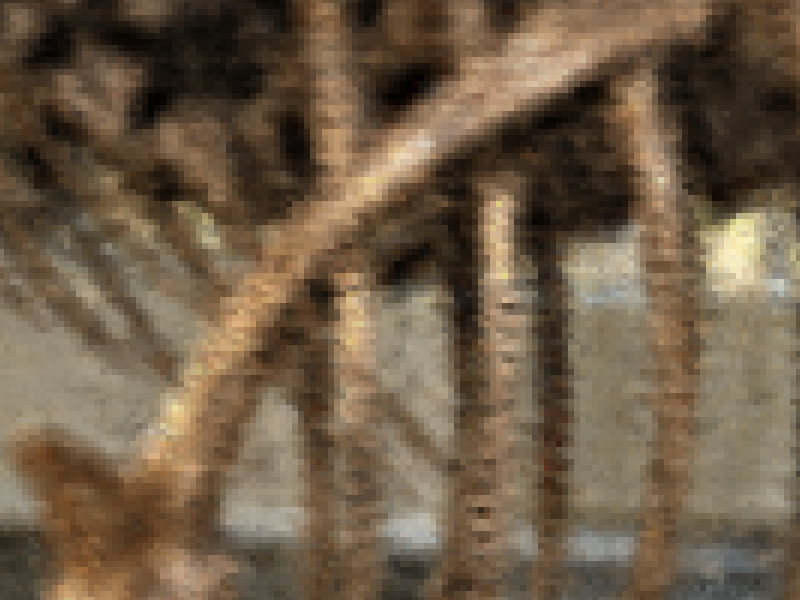}  &
\includegraphics[]{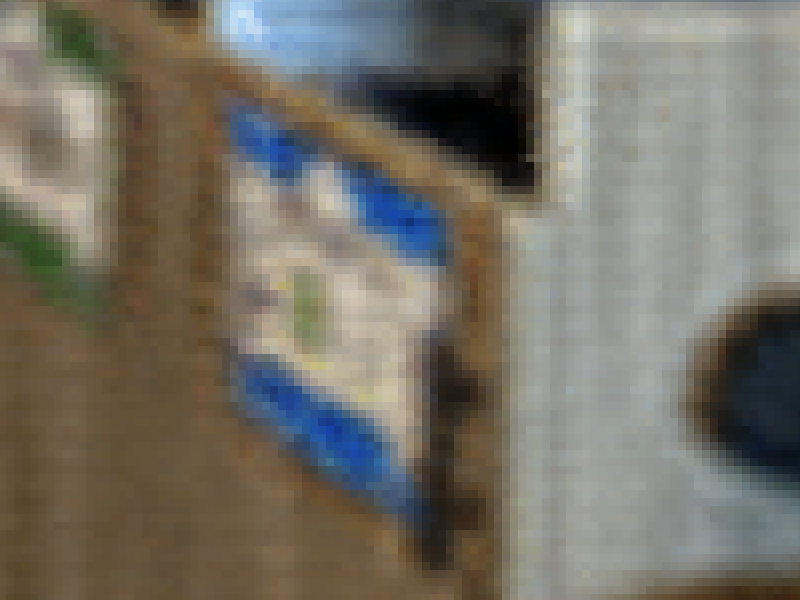}\\

\rotatebox{90}{\hspace{4cm}\fontsize{50}{200}\selectfont Ours}&
\includegraphics[]{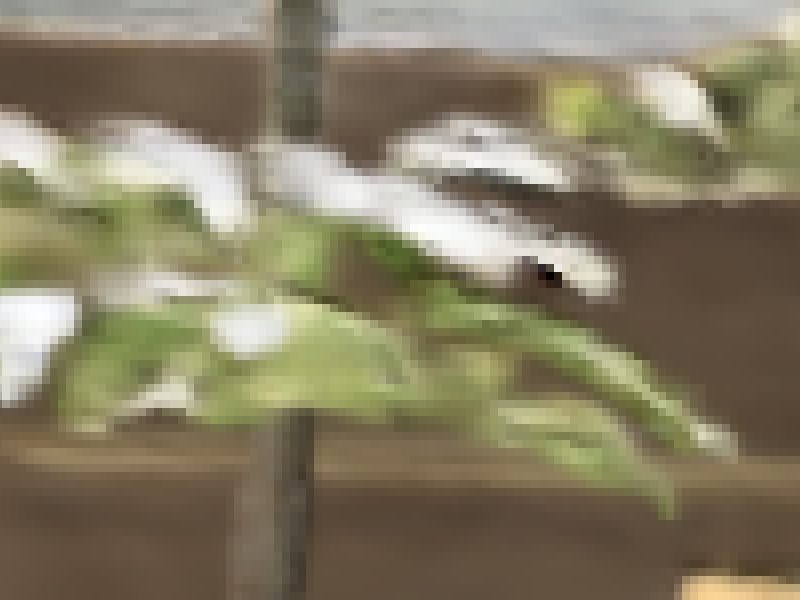} &
\includegraphics[]{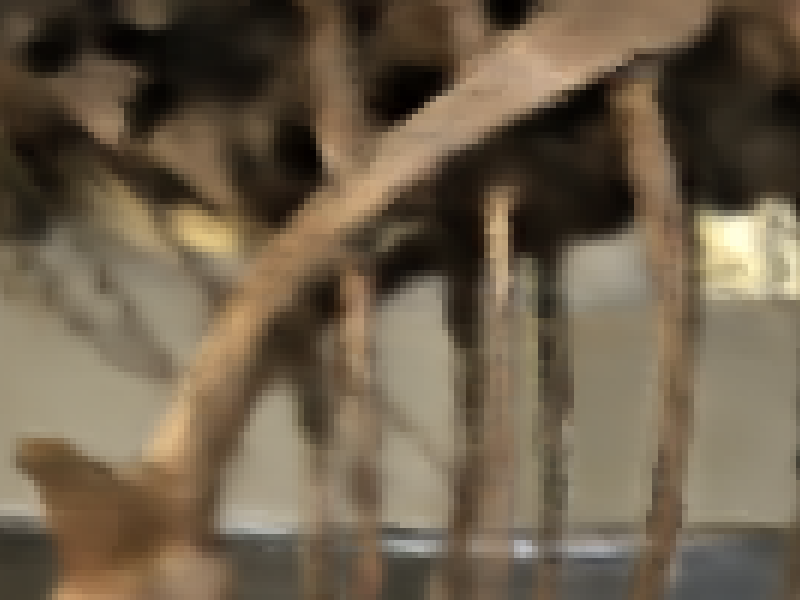} &
\includegraphics[]{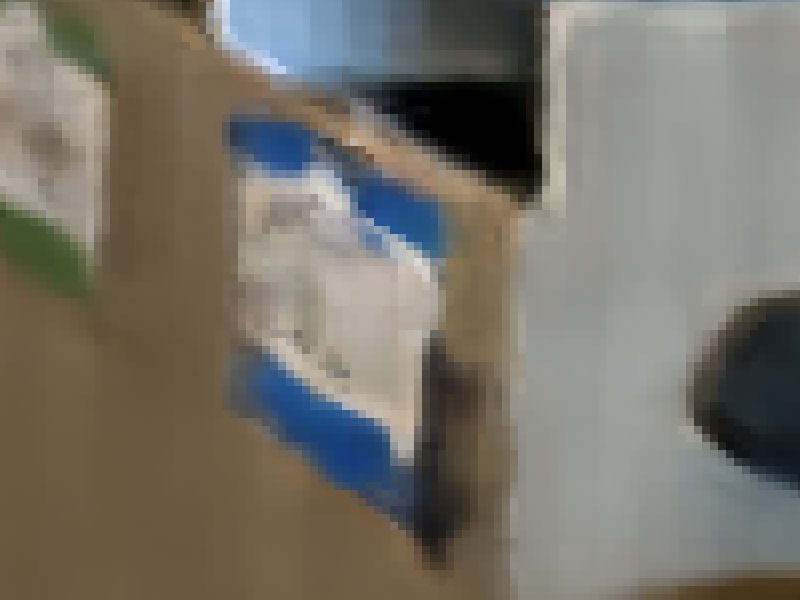} \\

\end{tabular}}

\caption{\textbf{Qualitative results on LLFF.} Our method renders more realistic HR images in scenes with complex backgrounds.}
\vspace{-4mm}
\label{fig:llff_qualitative}
\end{figure}

In main paper, we conduct experiments on datasets of bounded scenes. These scenes are object-centric and have simple backgrounds~\cite{mildenhall2020nerf,yao2020blendedmvs}. In this section, we provide results on LLFF~\cite{mildenhall2019locallightfieldfusion}, a dataset containing forward-facing scenes with complex objects and backgrounds. Following the same experiment settings from Sec.5.2. of the main paper, we compare our method with Zip-NeRF on LLFF. As shown in Tab.~\ref{tab:llff}, ASSR-NeRF outperforms Zip-NeRF with high upscaling factor (x4). Fig.~\ref{fig:llff_qualitative} also shows that our method can effectively improve the geometry and achieve cleaner appearances even when reconstructing scenes with complex backgrounds and objects.

\begin{table}[hbt!]
% \huge
\centering
\resizebox{0.5\linewidth}{!}{%
\begin{tabular}{c|ccc}
\toprule
x4&PSNR$\uparrow$&SSIM$\uparrow$&LPIPS$\downarrow$ \\ \midrule
Zip-NeRF	&23.351	&0.690	&0.419 \\
Ours(ASSR-NeRF)	&\textbf{23.801}	&\textbf{0.725}	&\textbf{0.361} \\
\bottomrule
\end{tabular}}
\vspace{5mm}
\caption{\textbf{Quantitative results on LLFF.} The experiment was trained on 252x189 image resolutions and tested on 1008x756.}
\label{tab:llff}
\end{table}

% \clearpage  % TODO REVIEW/FINAL: This \clearpage needs to be removed from both review and camera-ready versions.

% \bibliography{main}

\end{document}